# Singular points detection with semantic segmentation networks

Jiong Chen, Heng Zhao, Zhicheng Cao and Liaojun Pang*

## Abstract

Singular points detection is one of the most classical and important problem in the field of fingerprint recognition. However, current detection rates of singular points are still unsatisfactory, especially for low-quality fingerprints. Compared with traditional image processing-based detection methods, methods based on deep learning only need the original fingerprint image but not the fingerprint orientation field. In this paper, different from other detection methods based on deep learning, we treat singular points detection as a semantic segmentation problem and just use few data for training. Furthermore, we propose a new convolutional neural network called SinNet to extract the singular regions of interest and then use a blob detection method called SimpleBlobDetector to locate the singular points. The experiments are carried out on the test dataset from SPD2010, and the proposed method has much better performance than the other advanced methods in most aspects. Compared with the state-of-art algorithms in SPD2010, our method achieves an increase of 11% in the percentage of correctly detected fingerprints and an increase of more than 18% in the core detection rate.

*Keywords:* Singular points detection, semantic segmentation, blob detector, autoencoder

## Introduction

Fingerprint recognition was originally used for criminal investigation and has gradually extended to applications such as border control, computer logon, mobile payment [1], and so on. The singular point, which plays essential role in fingerprint recognition, fingerprint indexing and fingerprint template protection, is one of the most important and obvious global features for the fingerprint. The singular point area is defined as a region where the ridge curvature is higher than normal and where the direction of the ridge changes rapidly [3]. Fingerprint image processing and analysis are typically based on the location and pattern of singular points detected in the images. These singular points (cores and deltas) not only represent the characteristics of local ridge patterns but also determine the topological structure (i.e., fingerprint type) and largely influence the orientation field [2].

Singular points detection has been carried out for many years and there have been a lot of classic methods using traditional digital image processing technology, such as methods based on the Poincaré index[4, 5], filter technique[6-8] and others[9-14]. In recent years, the singular points detection based on deep learning have come up[15, 16], whose advantage over other methods is that the original fingerprint images were directly adopted instead of the orientation fields. Jin et al.[15] regarded the singular points detection as the block classification, while Liu et al.[16] regarded it as object detection using anchor-based detection method. However, they all used large

___

Jiong Chen, Heng Zhao, Zhicheng Cao and Liaojun Pang are with the Molecular and Neuroimaging Engineering Research Center of MOE, School of Life Science and Technology, Xidian University, Xi'an, China.
*Send correspondence to Liaojun Pang at e-mail: ljpang@mail.xidian.edu.cn.

amount of training data. For precise point detection, the block classification method was somewhat rough. The anchor-based object detection method was more precise, but the detection accuracy was still affected by anchor size[25]. Furthermore, the faster-RCNN(Faster Region-based Convolutional Network method) architecture for low-quality small target detection did not work very well[25]. Overall, for the state-of-the-art classic algorithms or those based on deep learning, the detection rate of singular points is still unsatisfactory, especially for the low-quality fingerprint.

In recent years, the deep learning methods in the pixels-to-pixels learning manner for semantic segmentation have been developing rapidly [17, 18, 21]. Compared to the block classification and object detection, semantic segmentation can predict each pixel. In this paper, different from the present singular points detection based on deep learning, we treat it as a semantic segmentation problem and design a semantic segmentation network architecture based on the encoder-decoder convolutional neural network with skip connections and the inception module. The new singular points detection network, called SinNet, only uses less training data for small singular regions segmentation and the blob detection method is used to locate the singular points. Our whole work consists of two parts: the first one is SinNet which is a pixels-to-pixels system to segment the singular point area, and the second one is the blob detection to locate the singular point precisely.

The contributions of this study are mainly three-fold:

1. Different from other deep learning methods, we treated singular points detection as a semantic segmentation problem. Besides, our method utilizes much less fingerprint data and did not need much data processing for training.

2. A new deep network architecture (SinNet) is designed for singular points detection. The application of the encoder-decoder structure, skip connections and the inception module ensures the excellent performance of the network.

3. Our method achieved the best accuracy of SPD2010. In detail, the performance of our method is much better than the other methods in most aspects, especially the correctly detection rate (CD) and the detection rate of cores. Compared with the state-of-art algorithm in SPD2010, our method achieves an increase of 11% in the percentage of correctly detected fingerprints and an increase of more than 18% in the core detection rate.

## Related works

Much research on singular points detection was based on traditional digital image processing algorithms, while deep learning, which has been used widely in the field of computer vision, has provided a new way for solving singular points detection. Therefore, the present singular points detection methods can be divided into two major categories: (1) ones on traditional digital image processing [4-14], and (2) others on deep learning [15, 16].

For the former methods, Poincare index [4] is the most classical way to detect singular points. Kawagoe et al. firstly came up with a judgement method of singular point according to the Poincaré index calculated for each point in the fingerprint's orientation field. On this basis, many researchers put forward many improvement. Lingling et al.[26] proposed an algorithm which combined Zero-pole Model and Hough Transform(HT) to detect the core and the delta. The orientation field was estimated by Zero-pole Model, HT was used for coarse detection at the global level and the Poincaré index was used for fine detection. Jie et al.[2] presented an ORIentation values along a Circle(DORIC) feature for the removal of spurious singular points. The classic Poincaré Index method was employed for detecting the singular points initially. As a novel constraint, a core-delta relation was utilized to select singular points finally.

Jin et al. [10] developed a new index of singular points based on the definition of Angle Matching Index (AMI) in vector fields. AMI was a specific polynomial model of orientation field, which could be also used in other areas, especially the vector field analysis. The AMI information of candidate singular points was collected and the conventional convergence index filter framework was modified. En et al. [20] proposed a faster method by walking directly to the singular points. The method did not need to visit each pixel or each small image block to locate the singular point. Walking Directional Fields (WDFs) should be established at first, and then a walking strategy was easily carried out for fast location with an acceptable accuracy. Most of the above traditional digital image processing methods needed the orientation fields in the processing of detection, meanwhile, they always did not work well for low-quality fingerprint images.

In recent years, the deep learning methods have appeared. Jin et al.[15] firstly proposed the singular points detection method based on convolutional networks. They designed two convolutional neural networks. One was used for classifying the core, the delta and the background and the other was used for estimating the positions of the core and the delta. At last, a probabilistic method was used to determine the actual positions of singular points. However, the block classification method is somewhat rough. And this method had many processing steps. Its training process was also cumbersome including dividing the training images into many blocks and multi-scales. Liu et al. [16] was inspired by the successful object detection in natural images, and regarded the singularity detection as the object detection problem and proposed an deep learning detection algorithm via faster-RCNN. This algorithm had a two-step strategy. Some candidate patches which had probabilities of existing singular points were chosen by one network in the first step. The precise singular points were selected and located from the candidate patches in the second step. Compared to the Jin's method, this algorithm based on the anchor-based object detection is more precise. However, the detection accuracy is still affected by anchor size, and low-quality small target detection is still a difficult problem [25]. The faster-RCNN does not work very well for this problem. At the same time, it needed a large number of fingerprint images for training. Actually, the actual experimental results of these methods based on deep learning were not satisfactory, especially for low-quality fingerprint images.

## Proposed method

Convnets are not only improving for whole-image classification, but also making progress on object detection, part and key point prediction, and local correspondence. The natural next step in the progression from coarse to fine inference is to make a prediction at every pixel which is called semantic segmentation [21]. Inspired by the definition of the singular point area and the processing ways of some classic singular methods, we reasonably believe that it is suitable to regard the singular point area detection as a semantic segmentation problem. We try to detect singular point in two steps: at first, finding the singular region of interest (SROI), and then locating the accurate singular point.

For generic semantic segmentation task, convolutional neural network especially encoder-decoder architecture network usually needs to segment more than a dozen objects of different subject classes in natural images. Different objects in an image usually contain different features, such as texture, color, shape and size. Compared to them, the fingerprint image is simpler. Fingerprint images are usually gray scale images. Their ridges are represented by dark lines and their valleys together with the background are represented by bright pixels. To some extent, the structures of singular points are simpler than generic objects so that we adopt a shallow encoder-decoder architecture network to achieve the singular point area segmentation task. At the same time, in order to enhance robustness for cores and deltas of different size, we design a wide network using inception module [22] for extracting multi-scale features. However, different from the generic semantic segmentation task, there are a lot of low-quality fingerprint images in the task of the singular point area segmentation. Inspired by the researches on the super resolution [27][28], we adopt the symmetrical structure and the skip connection to restore the details of fingerprint images.

The flowchart of the proposed method is shown in Fig.1. Our detection process consists of two parts: the first part is the SinNet for singular region of interest (SROI) which has dual channel structure, one branch for the core area detection and the other branch for the delta; the second part is a blob detector for accurate singular point location. All 210 fingerprint images from SPD2010 training dataset are adopted to train the SinNet.

3.1.1 Training data preparation

Compared to other deep learning methods, our method utilizes much less fingerprint data for training and does not need complex data processing. The training set we use is the training set of SPD2010. SPD2010 was the first fingerprint singular points detection competition, which has marked singular point coordinates. SPD2010 benchmark database has 500 fingerprint images with 355×390 pixel resolution captured by an optical scanner (Microsoft Fingerprint Reader - model 1033) without any restrictions on the poses of fingers. The subjects are males and females aged from 20 to 62 years old coming from 7 countries. The fingerprint images in this dataset have a large variety in quality, type, affine transformation and nonlinear distortion [23]. The ground truth for the positions of core and delta points is obtained by hand according to E. R. Henry's definition [11] of singular points. In the course of training, any data augmentation was not used. Fig.2 shows some examples from the training set of SPD2010. Fig. 1 shows three examples of the training set of SPD2010.

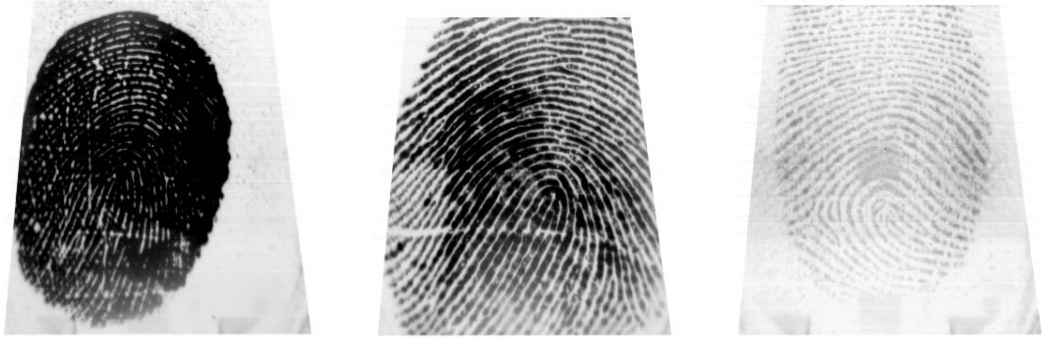

Fig. 1: Three examples of the training set of SPD2010.

### 3.1.2 Ground truth labels

According to SPD2010 competition's instructions, for a ground truth singular point **(x0; y0; t0)**, if a detected singular point **(x;y;t)**, satisfies **(t=t0)** and **sqrt((x-x0)^2+(y-y0)^2)<10**, it is said to be truly detected. This circular area is considered to be the singular point area. In the process of labeling, the singular point coordinates as the center and 10 pixels as the radius were set to draw a circle. The circle was set to the foreground and the other areas were set to the background. Our labels are classified into two categories, and two sets of labels are made. One set is the label of cores and the other is the label of deltas. The calibrated samples of label are shown in fig. 2:

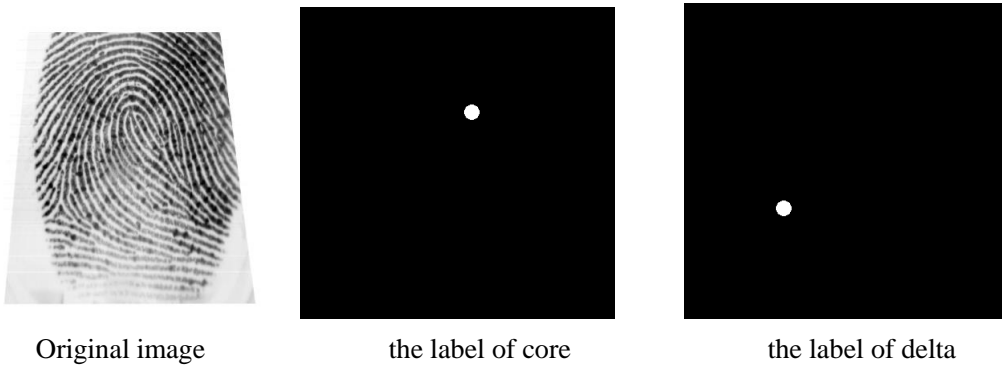

Original image          the label of core          the label of delta

Fig. 2: the calibrated samples of label

### 3.2 Network architecture

Like many convolutional neural networks for semantic segmentation, the encoder-decoder architecture is adopted in our SinNet. The function of encoder part is feature extraction. In the convolution neural network, inception module and skip connection are used. There is the the rectified linear units(ReLU) operation after each convolutional layer. The decoder part is to up-sample the abstraction back into its original size in detail. In order to prevent over-fitting, we also adopted dropout. The structure of SinNet is shown in the Fig. 3. And the details of the network are introduced in Section 3.2.4.

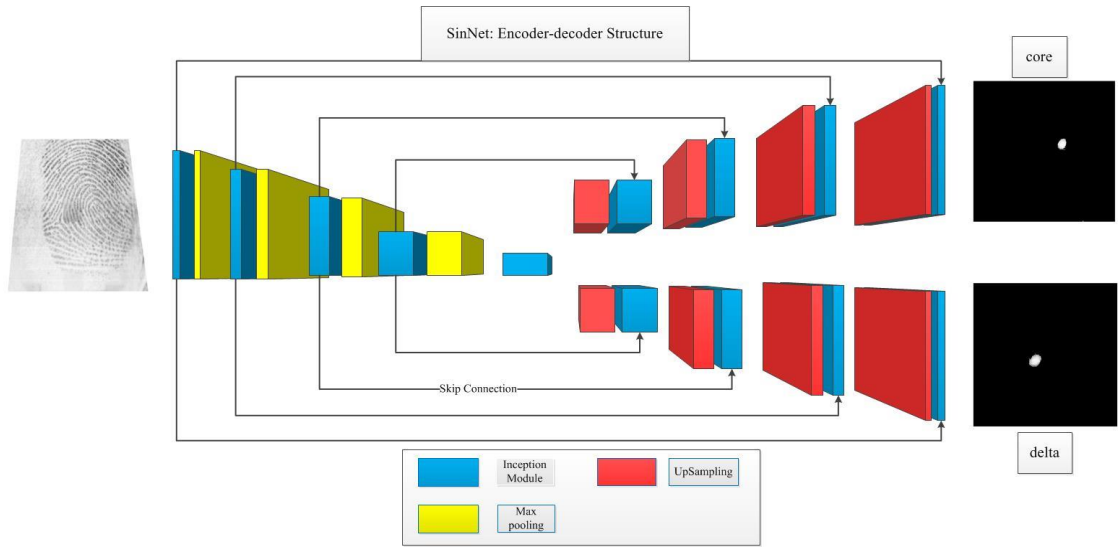

Fig. 3: the structure of SinNet

3.2.1 Encoder-decoder architecture
The encoder-decoder architecture is inspired by the autoencoder. The autoencoder is a kind of neural network. After training, it can try to copy input to output. At first, it was used for data dimension reduction and data generation. Semantic segmentation faces an inherent tension between semantics and location: global information resolves what while local information resolves where[21]. The encoder-decoder architecture is widely applied for the semantic pixel-wise segmentation, such as FCN, U-net, SegNet etc. The encoder layers act as a feature extractor, which aims at extracting target characteristics. The decoder layers are combined to recover the details of semantic segmentation. The architecture has been proved to work well for semantic segmentation. In this paper, the symmetric encoder-decoder architecture is adopted for dense skip connections conveniently.

3.2.2 Inception module
Inception network is an important milestone in the development of CNN classifier. Before Inception network, most popular CNNs just stacked more and more convolution layers, making the network deeper and deeper, hoping to get better performance. The greatest feature of GoogLeNet was the use of inception module. Its purpose is to design a network with excellent local topology, that is, to perform multiple convolution or pooling operations on input images in parallel, and to stitch all output results into a very deep feature map. Inception module is a local scaling, which can help the network extract features on different scales, increase different field of perception. Inspired by Inception V3 [22], in the whole SinNet we adopted a lot of inception modules, and the modules have the same structure. The structure of this module is shown in the Fig. 4 below.

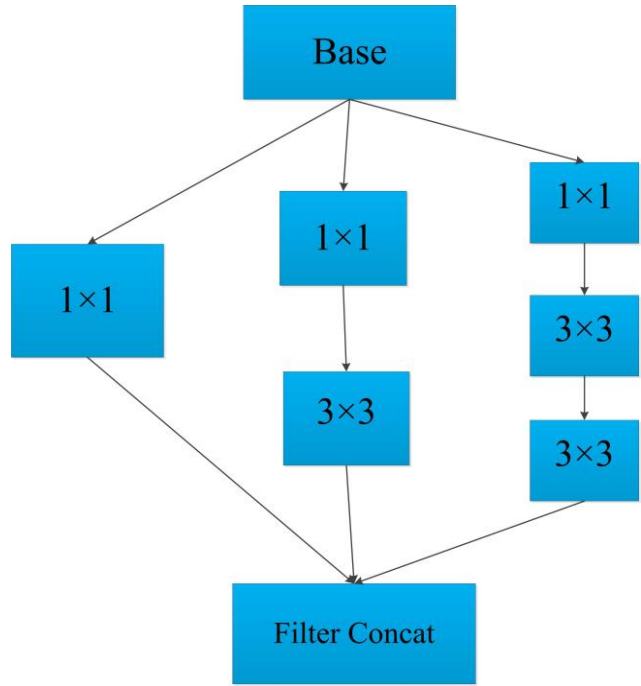

Fig. 4: inception module of SinNet

3.2.3 Skip connection

The symmetric encoder-decoder architecture is adopted in the SinNet. For the SinNet, the encoder layers including down-sampling act as a feature extractor. However, with the continuous convolution and down-sampling, the receptive field becomes larger and larger, and the semantics becomes more and more advanced, but the low-level information will also be lost. To solve this problem, skip connections are occurring between two symmetrical layers. The use of skip connections ensures that the final recovered feature map fuses low-level features. Moreover, features of different scales can be fused.

3.2.4 Details of SinNet

The SinNet architecture is illustrated in Fig. 3. The SinNet consists of two parts. The upper part is the encoder part and the lower part is the decoder part. The upper part is public and the lower part has two branches. One branch is for the core area segmentation, and the other branch is to segment the delta area. The upper part has five sub-modules. In the SinNet, the inception module inspired by Inception V3 [22] is designed as Fig.4. From the first to the forth sub-module, those four sections are all composed of one inception module and one 2*2 max pooling layer. Their filter sizes are corresponding, but the numbers of filters are different. From the first to the forth sub-module, the numbers of filters are 64, 128, 256 and 512 respectively. For the fifth sub-module, it is just one inception module, whose structure is same with the upper ones and the number of filter is 1024. For the lower part, the two branches are identical in structure and different in function. One branch is for the core area segmentation, and the other is for the delta area segmentation. One branch consists of four sub-modules. From the first to the third sub-module, those three sections are all composed of one 2*2 up-sampling layer and one inception module. Their filter sizes are corresponding, but the numbers of filters are different. From the first to the third sub-module, the numbers of filters are 512, 256, 128 respectively. For the forth sub-module, it consists one 2*2 up-sampling layer, one inception module and two convolution layers. The number of filter of inception module is 64, and the numbers of filters of two convolution layers are 2 and 1 respectively. Actually, the last convolution layer is the output. The output is a gray image, and the bright pixels mean that there may be singularities, while the dark pixels mean that there may be the background. The output is shown in Fig. 3.

3.2.5 Training

We use the cross-entropy loss [2] as the objective function for training the network. The loss is summed up over all the pixels in a mini-batch. The optimization method is random gradient descent (SGD). The learning rate is set at 0.1. The momentum is set at 0.9. The mini-batch is set at 16. The epoch is set at 100.

3.3 Singular point localization based on blob detection

After the trained network SinNet processing, the test fingerprint image from SPD2010 can get a binary image, and the response area is the possible singular point area. Our purpose is to locate the precise location of the singular point. In this paper, the blob detection is adopted and the center of the detected blob is regarded as the singular point.

A blob is a group of connected pixels in an image that share some common property. The singular point regions we want to identify and extract are typical blobs. Blob detection mainly detects the pixel areas whose gray value is larger or smaller than the surrounding area in the image. Common methods are based on differential detector and local extremum. OpenCV, a computer vision library, provides a blob detection method called SimpleBlobDetector, which can be controlled by multiple parameters. The algorithm is simple and fast[24].

Many parameters in SimpleBlobDetector need to be set which can limit the size, shape, color and so on of the blob to be detected. For the the possible singularity area, there are two important parameters. One is the MinArea, which is for removing noise, because too small response area is generally not singular point area. The other one is the MaxArea. This value cannot be set too large because the center of large response area is generally not precise enough so that it always results in False Alarm. In our experiment, the MinArea is set at 100 and the MaxArea is set at 800. The example of blob detection result is shown in Fig. 5. The first star represents the core, and the second star represents the delta.

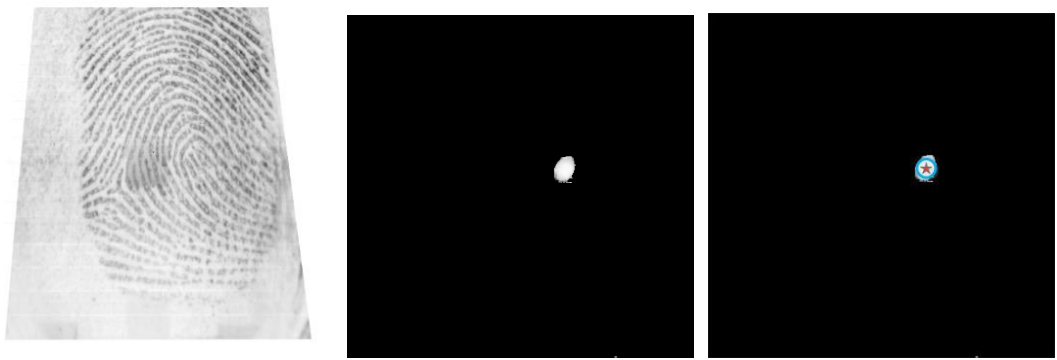

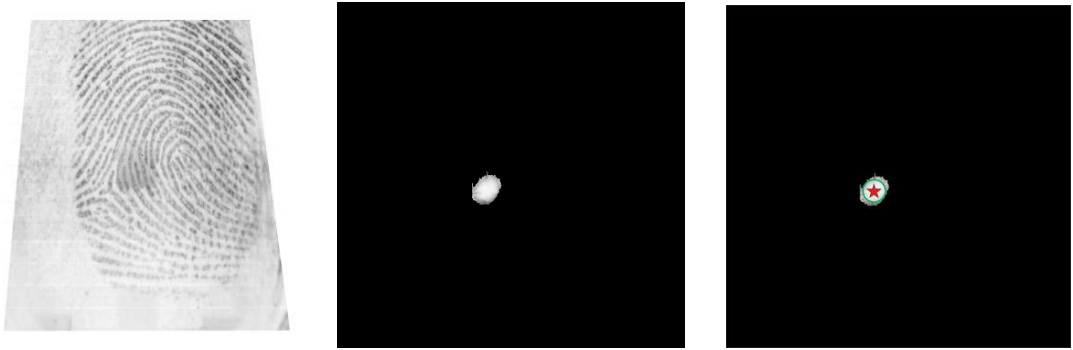

Fig.5: An example of blob detection result

After SimpleBlobDetector processing, the exact location and number of singular points are obtained from the fingerprint image. For a fingerprint, the number of cores is at most two, and the number of deltas is at most two. If a fingerprint image detects three or more cores or deltas, to reduce the false alarm, the detection result is considered that there are no cores or deltas.

## Experimental results

In this section, we show the experimental results and analysis conducted on SPD2010 test dataset. First, the testing fingerprints datasets in this work and evaluation criteria are introduced. Then, the experimental results and analysis are presented. We also compared with other advanced algorithms in the literature to show the effectiveness of our final results.

4.1. Datasets and parameters

The proposed algorithm is tested on the first fingerprint singular points detection competition (SPD2010). In our work, the training data of SPD2010 is used for training and the testing data of SPD2010 is used for testing. According to SPD2010, the evaluation criteria are as follows: the evaluation will be based on measures that will consider the quantity and type of the detected singular points as well as their distance to the ground truth, which is manually labeled beforehand. For a ground truth singular point, **(x0;y0;t0)**, if a detected singular point **(x;y;t)**, satisfies **(t=t0)** and **sqrt((x-x0)^2+(y-y0)^2)<10**, it is said to be truly detected and, otherwise, it is called a miss. The detection rate is defined as the ratio of truly detected singular points to all ground truth singular points. The miss rate is defined as the ratio of the number of missed singular points to the number of all ground truth singular points. The false alarm rate is defined as the number of falsely detected singular points versus the number of all ground truth singular points. If all singular points are detected and there are no spurious singular points in a fingerprint, the fingerprint is considered to be 'correctly' detected [23].

4.2. Experimental analysis and comparison with existing state-of-the-art methods

Our method is compared with three state-of-the-art methods: angle matching index combined with convergence index filter (AMF, also a model-based method) method [19], walking point method [20] and Faster-RCNN [16] method. All the above methods are evaluated according to the evaluation criteria of SPD2010 competition. The training set is used for training and the testing set is used for testing.

Table 1 the performance of the three singular points detection methods and the proposed method over the SPD2010 test dataset

| Algorithm | CD(%) | Detection rate (%) | | Miss rate (%) | | False alarm rate (%) | |
|---|---|---|---|---|---|---|---|
| | | Cores | deltas | Cores | deltas | Cores | deltas |
| AMF[19] | 38 | 52 | 60 | 48 | 40 | 56 | 28 |
| Walking point [20] | 32 | 42 | 53 | 58 | 47 | **10** | **12** |
| Faster-RCNN[16] | 37 | 56 | 48 | 44 | 52 | 30 | 35 |
| Proposed Method | **49** | **70** | **61** | **30** | **39** | 15 | 18 |

The values of the above metrics for test data are listed in Tab 1. Our algorithm's performance achieves the state-of-the-art accuracy. Actually, the performance of our method is much better than the other top 3 competitors in the SPD2010 test dataset in most aspects, especially the correctly detection rate (CD) and the detection rate of cores. For example, compared with the best traditional algorithm (called AMF) in SPD2010, our method achieves an increase of 11% in the percentage of correctly detected fingerprints and an increase of more than 18% in the core detection rate in the test dataset from SPD2010. We also compare our algorithm with the state-of-the-art deep learning algorithm Faster-RCNN method on the same benchmark. Our method achieves an increase of 12% in the percentage of correctly detected fingerprints and an increase of more than 14% in the core detection rate in the test dataset from SPD2010. The walking point worked well in the aspect of False Alarm, but as can be seen from the table I, it did much worse in other aspects, especially the most import correctly detection rate (CD).

In Fig. 6, we present the detection results obtained by different methods on two low-quality fingerprint examples in the test dataset from SPD2010. The subgraph (a) and (d) show the core detection results of No.0311 and No.0436 in the test dataset. The red star is the ground truth, the green star is the result obtained by the proposed method, the yellow star is the result obtained by Faster-RCNN, the blue star is the result obtained by AMF and the black circle means the corrected detection area. The subgraph (b), (c), (e), and (f) show the details of the core detection results obtained by the proposed method. As can be seen from the Fig. 6, the performance of our method is good, while AMF and Faster-RCNN detect error singular points, and the walking point cannot even detect any point. It can be proved that our method is more effective and robust to detect the singular points of the low-quality fingerprint.

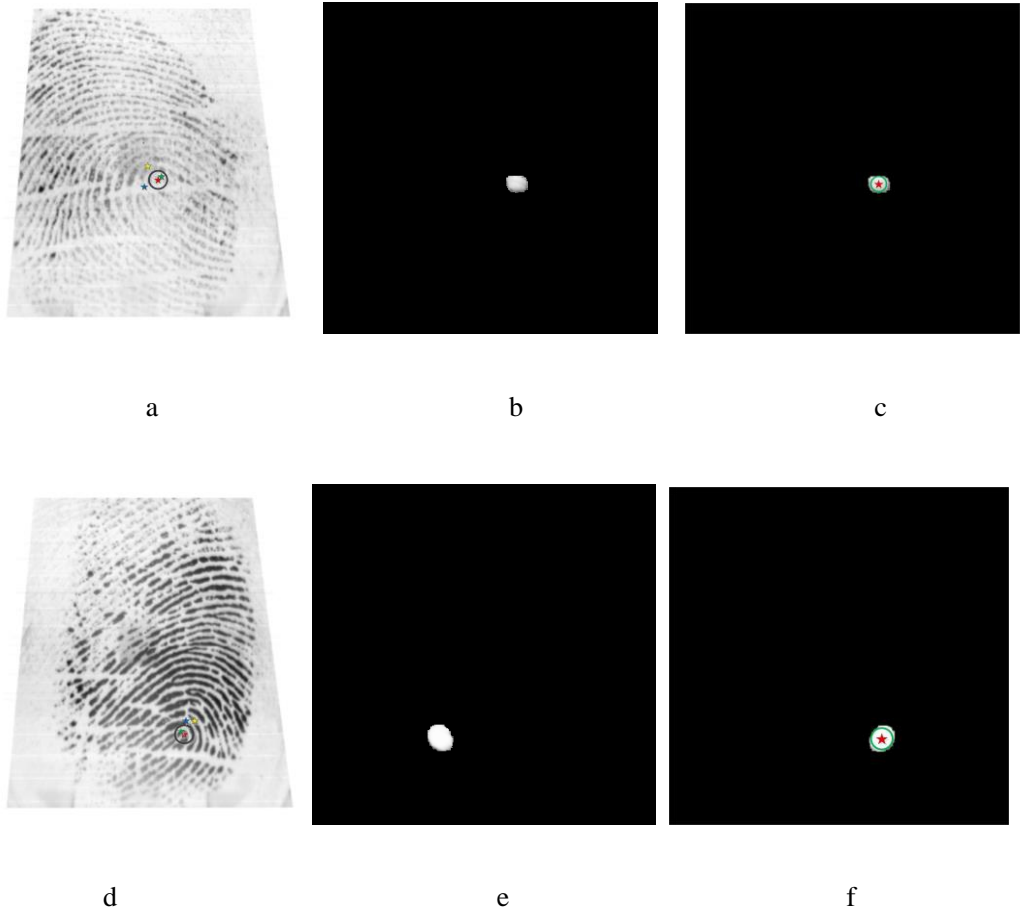

| a | b | c |
| d | e | f |

Fig. 6: the core detection results obtained by different methods on the low-quality fingerprints

## Conclusions

In this paper, we proposed a new convolutional neural network called SinNet to extract the singular regions of interest and then used the SimpleBlobDetector to locate the singular points. The SinNet is a pixels-to-pixels system. First of all, different from other methods based on deep learning, we treated singular points detection as a semantic segmentation problem in a new respective and the strategy worked very well. After we have analyzed the similarities and differences between common semantics segmentation problems and the singular point detection carefully, a new deep network architecture (SinNet) is designed. In addition to the encoder-decoder structure, the SinNet consists of the skip connection and the inception module. Secondly, our method utilizes much less fingerprint data and does not need much data processing. Thirdly, in the test dataset from SPD2010, our method designed and trained for the task of singular points detection achieves the state-of-the-art accuracy. Actually, the performance of our method is much better than the other advanced methods in most aspects, especially the correctly detection rate (CD) and the detection rate of cores.

In our future work, we will improve the labeling process. In this process of labeling, the singular point coordinates as the center and 10 pixels as the radius were set to draw a circle. The circle was set to the foreground and the other area was set to the background. The segmentation of the singular region can be more precise. Furthermore, we will design new convolutional neural network on the basis of SinNet to improve the detection accuracy, especially the detection rate of deltas.

## Acknowledgement

This research is funded by the National Cryptography Development Fund under Grant MMJJ20170208, and the National Natural Science Foundation of China under Grants 61876139 and 61906149.

## References

[1] Gu S , Feng J , Lu J , et al. Efficient Rectification of Distorted Fingerprints[J]. IEEE Transactions on Information Forensics and Security, 2017, PP(99):1-1.
[2] Zhou J , Chen F , Gu J . A novel algorithm for detecting singular points from fingerprint images.[J]. IEEE Transactions on Pattern Analysis & Machine Intelligence, 2009, 31(7):1239.
[3] F. Magalhaes, H. P. Oliveira, and A. Campilho. Spd 2010 fingerprint singular points detection competition introduction. [Online]. Available: http://paginas.fe.up.pt/ ∼ spd2010/
[4] D. Weng, Y. Yin, D. Yang, Singular points detection based on multi-resolution in fingerprint images, Neurocomputing 74 (17) (2011) 3376–3388.
[5] F. Belhadj, S. Akrouf, S. Harous, S.A. Aoudia, Efficient fingerprint singular points detection algorithm using orientation-deviation features, J. Electron. Imaging 24 (3) (2015) 033016–033016.
[6] K. Nilsson, J. Bigun, Prominent symmetry points as landmarks in fingerprint images for alignment, in: Proceedings of the 16th International Conference on Pattern Recognition (ICPR), vol. 3, IEEE, 2002, pp. 395–398.
[7] K. Nilsson, J. Bigun, Localization of corresponding points in fingerprints by complex filtering, Pattern Recognit. Lett. 24 (13) (2003) 2135–2144.
[8] H. Fronthaler, K. Kollreider, J. Bigun, Local feature extraction in fingerprints by complex filtering, in: Advances in Biometric Person Authentication, Springer, Berlin Heidelberg, 2005, pp. 77–84.
[9] L. Fan, S. Wang, H. Wang, T. Guo, Singular points detection based on zero-pole model in fingerprint images, IEEE Trans. Pattern Anal. Mach. Intell. 30 (6) (2008) 929–940.
[10] J. Qi, S. Liu, A robust approach for singular point extraction based on complex polynomial model, in: Proceedings of the 2014 IEEE Conference on Computer Vision and Pattern Recognition Workshops (CVPRW), IEEE, 2014, pp. 78–83.
[11] Henry E R . Classification and Uses of Finger Prints[J]. Classification & Uses of Finger Prints, 1934.
[12] C.-H. Park, J.-J. Lee, M.J. Smith, K.-H. Park, Singular point detection by shape analysis of directional fields in fingerprints, Pattern Recognit. 39 (5) (2006)839–855.
[13] C. Jin, H. Kim, Pixel-level singular point detection from multi-scale Gaussian filtered orientation field, Pattern Recognit. 43 (11) (2010) 3879–3890.
[14] H. Chen, L. Pang, J. Liang, E. Liu, J. Tian, Fingerprint singular point detection based on multiple-scale orientation entropy, IEEE Signal Process. Lett. 18 (11) (2011) 679–682.
[15] Qin, J.; Han, C.; Bai, C.; Guo, T. Multi-scaling detection of singular points based on fully convolutional networks in fingerprint images. In Proceedings of the Chinese Conference on Biometric Recognition (CCBR), Shenzhen, China, 28–29 October 2017; pp. 221–230.

[16] Liu Y, Zhou B, Han C, et al. A Method for Singular Points Detection Based on Faster-RCNN[J]. Applied Sciences, 2018, 8(10): 1853.

[17] Ronneberger O, Fischer P, Brox T. U-net: Convolutional networks for biomedical image segmentation[C]//International Conference on Medical image computing and computer-assisted intervention. Springer, Cham, 2015: 234-241.

[18] Badrinarayanan V, Kendall A, Cipolla R. Segnet: A deep convolutional encoder-decoder architecture for image segmentation[J]. IEEE transactions on pattern analysis and machine intelligence, 2017, 39(12): 2481-2495.

[19] Qi J, Liu S. A robust approach for singular point extraction based on complex polynomial model[C]//Proceedings of the IEEE Conference on Computer Vision and Pattern Recognition Workshops. 2014: 78-83.

[20] Zhu E, Guo X, Yin J. Walking to singular points of fingerprints[J]. Pattern Recognition, 2016, 56: 116-128.

[21] Long J, Shelhamer E, Darrell T. Fully convolutional networks for semantic segmentation[C]//Proceedings of the IEEE conference on computer vision and pattern recognition. 2015: 3431-3440.

[22] Szegedy C, Vanhoucke V, Ioffe S, et al. Rethinking the inception architecture for computer vision[C]//Proceedings of the IEEE conference on computer vision and pattern recognition. 2016: 2818-2826.

[23] https://paginas.fe.up.pt/~spd2010/

[24] Bradski G, Kaehler A．Learning OpenCV：Computer vision with the OpenCV library, [M]．Sebastopol：O'Reilly Media，Inc. 2008．

[25] Kisantal M , Wojna Z , Murawski J , et al. Augmentation for small object detection[J]. 2019.

[26] L. Fan, S. Wang, H. Wang, T. Guo, Singular points detection based on zero-pole model in fingerprint images, IEEE Trans. Pattern Anal. Mach. Intell. 30 (6) (2008) 929–940.

[27] Mao X J, Shen C, Yang Y B. Image restoration using convolutional auto-encoders with symmetric skip connections[J]. arXiv preprint arXiv:1606.08921, 2016.

[28] Tong T, Li G, Liu X, et al. Image super-resolution using dense skip connections[C]//Proceedings of the IEEE International Conference on Computer Vision. 2017: 4799-4807.